\documentclass[10pt,twocolumn,letterpaper]{article}

\usepackage{iccv}
\usepackage{authblk}
\usepackage{times}
\usepackage{epsfig}
\usepackage{graphicx}
\usepackage{amsmath}
\usepackage{amssymb}
\usepackage{enumitem}
\usepackage{multirow}
\usepackage{booktabs}
\usepackage{cite}
\usepackage[T1]{fontenc}
\usepackage[utf8]{inputenc}
\usepackage[english]{babel}
\usepackage[autostyle, english=american]{csquotes}
\MakeOuterQuote{"}

\usepackage[breaklinks=true,bookmarks=false]{hyperref}

\iccvfinalcopy 

\def\iccvPaperID{****} 

\ificcvfinal\fi

\begin{document}

\title{The Solution for Single Object Tracking Task of  Perception Test Challenge 2024 }


\author{
  Zhiqiang Zhong$^1$,
  Yang Yang\textsuperscript{1, $\thanks{Corresponding Author}$},
  Fengqiang Wan$^1$,
  Henglu Wei$^2$,
  Xiangyang Ji$^2$,
}

\affil{
  $^1$Nanjing University of Science and Technology\\
  $^2$Tsinghua University\\
}

\maketitle
\ificcvfinal\thispagestyle{empty}\fi

\begin{abstract}
\quad This report presents our method for Single Object Tracking (SOT), which aims to track a specified object throughout a video sequence. We employ the LoRAT method. The essence of the work lies in adapting LoRA, a technique that fine-tunes a small subset of model parameters without adding inference latency, to the domain of visual tracking. We train our model using the extensive LaSOT and GOT-10k datasets, which provide a solid foundation for robust performance. Additionally, we implement the alpha-refine technique for post-processing the bounding box outputs. Although the alpha-refine method does not yield the anticipated results, our overall approach achieves a score of 0.813, securing first place in the competition.
\end{abstract}

\maketitle

\section{Introduction.}
\textbf{Team Introduction.} We have established ourselves as a hub for innovative studies in machine learning, data mining, and pattern recognition~\cite{DBLP:conf/mm/0074ZGGZ22,DBLP:journals/titb/ZhangYYGLZYR22,DBLP:conf/aaai/YangHGXX23,DBLP:journals/corr/abs-2401-07551,DBLP:journals/corr/abs-2404-08347}. Our primary research interests encompass foundational, cutting-edge, and transformative approaches aimed at advancing these disciplines.

We focus on several key areas, including multimodal machine learning, open-environment machine learning, and incremental learning. Additionally, we are dedicated to practical applications, exploring real-world scenarios such as image retrieval, video segmentation, and novel class detection~\cite{DBLP:conf/mm/0074ZGGZ22,DBLP:journals/tkde/YangYBZZGXY23,DBLP:conf/kdd/YangZZX019,DBLP:journals/titb/ZhangYYGLZYR22}. Our commitment to both theoretical advancements and their implementation in diverse contexts positions us at the forefront of knowledge-mining research, fostering an environment where collaboration and innovation can thrive~\cite{DBLP:journals/fcsc/YangGLLLY24,DBLP:conf/ijcai/YangZXYZY21,DBLP:conf/pakdd/Zhou0Z21,DBLP:conf/aaai/YangHGXX23}.

\textbf{Competition Introduction.} Benchmarking models is crucial for selecting the most optimal model and determining the most suitable development path for tracking tasks. While existing benchmarks have driven significant advancements in computational performance, many primarily focus on computational aspects rather than perceptual capabilities.

To address this gap, DeepMind has introduced the Perception Test, a novel multimodal video benchmark designed to evaluate the capabilities of multimodal perception models more comprehensively. Unlike traditional benchmarks, the Perception Test emphasizes perceptual tasks by leveraging real-world video data that are purposefully designed, filmed, and annotated. This benchmark aims to assess a wide range of skills, types of reasoning, and modalities in multimodal perception models~\cite{patraucean2024perception}. The Perception Test features an extensive dataset comprising 11,600 real-world videos, each with an average length of 23 seconds. These videos capture perceptually interesting scenarios and are filmed by approximately 100 participants from around the globe. The dataset is densely annotated with six types of labels, including multiple-choice questions, grounded video question-answers, object and point tracks, temporal action segments, and sound segments. This rich annotation allows for a thorough evaluation of both language-based and non-language-based models, providing a more holistic view of model performance across various perceptual tasks.

\begin{figure}[h]
 \centering
 \includegraphics[width=0.47\textwidth]{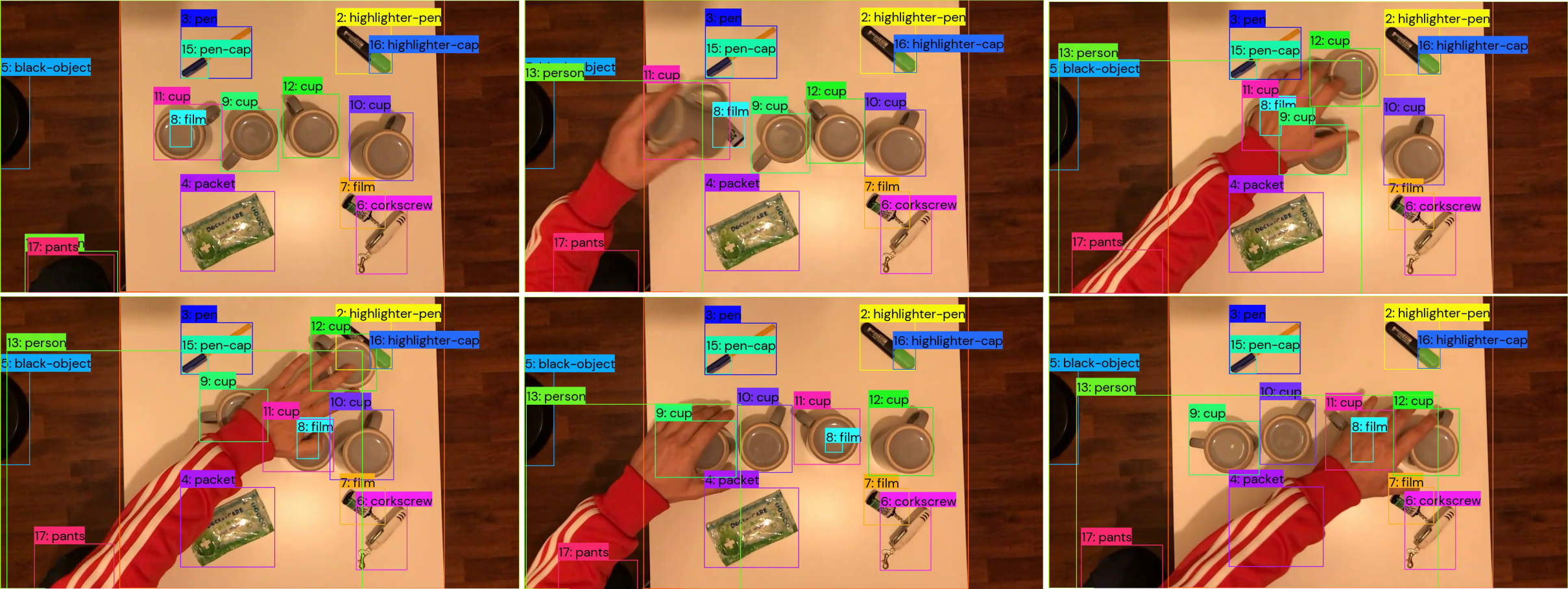}
 \caption{the examples of object tracking annotations.}
 \label{fig1}
\end{figure}

In the Perception Test Challenge 2024, our team selects single-object tracking. In this task, the model receives a video and a bounding box representing an object, and it tracks the object throughout the video sequence. Figure \ref{fig1} shows examples of object tracking annotations. We evaluate several prominent SOT models, including SiamFC~\cite{bertinetto2016fully}, SAM2~\cite{ravi2024sam}, and LoRAT~\cite{lin2024tracking}. After a thorough assessment, we choose LoRAT as the base model for our solution due to its superior performance and alignment with our tracking objectives. To enhance the model's capabilities, we incorporate additional strategies during the training phase.

We utilize supplementary datasets such as LaSOT~\cite{fan2019lasot} and GOT-10k~\cite{huang2019got} to broaden the model's exposure and improve its generalization across different scenarios. This expanded training dataset significantly contributes to the model's robustness and adaptability. Additionally, we implement the Alpha-Refine method to post-process the bounding box output generated by the model. Although Alpha-Refine aims to improve tracking performance through accurate bounding box estimation, its performance is not good enough. 

Despite the limited effect of the Alpha-Refine method, our approach, combining diverse training data and advanced tracking techniques, leads to significant performance improvements. These efforts ultimately result in our team achieving first place in the test phase of this track.

\begin{figure*}[h]
 \centering
 \includegraphics[width=\textwidth]{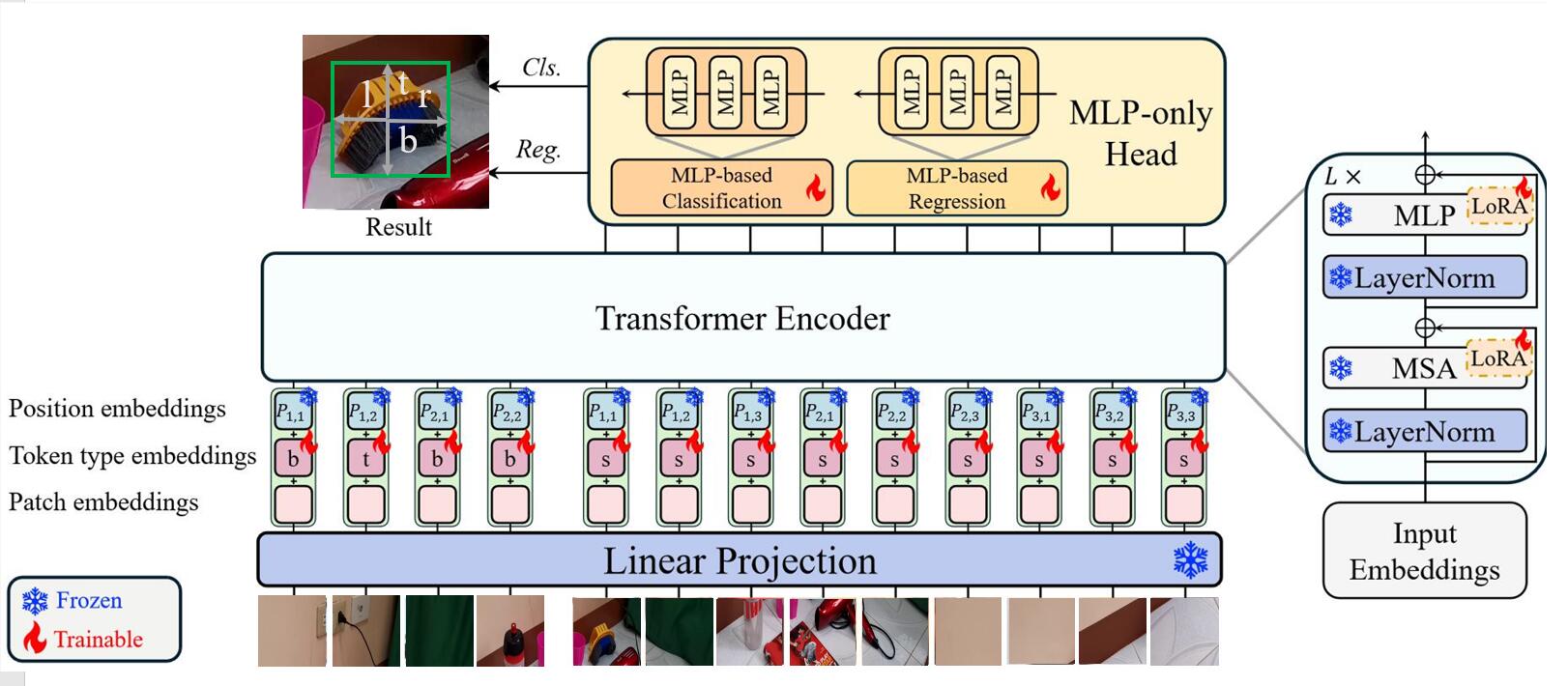}
 \caption{Architecture of LoRAT. }
 \label{fig2}
\end{figure*}

\section{Method}
This section provides an overview of the LoRAT approach~\cite{DBLP:journals/corr/abs-2403-05231}, highlighting its architectural components. Key elements include adjustments to the model’s input embedding for enhanced adaptability and integrating an MLP-only head network. These modifications are crucial for optimizing performance while minimizing computational overhead. Figure \ref{fig2} illustrates an overview of the tracker.

\subsection{LoRA-friendly Model Design}

\textbf{Decoupled Input Embedding.} ViT models require positional encodings to inject positional information~\cite{vaswani2017attention,yang2023towards}. Existing one-stream trackers utilize separate positional embeddings for template and search region tokens, which disrupts the structure of the pre-trained ViT model and leads to ineffectiveness in parameter-efficient fine-tuning (PEFT). To address this issue, adaptations for multi-resolution positional embeddings are explored, along with the integration of token-type embeddings.

\textit{Token type embedding.} Token type embedding, also known as segment embedding, is incorporated into the tracker by assigning unique embeddings to template and search region tokens. This approach decouples token type identification from positional embeddings, thereby enhancing parameter efficiency and flexibility.

The use of token-type embedding is further extended to address specific challenges associated with the cropping strategy commonly employed in Siamese trackers. This cropping strategy, used to generate tracking templates, presents several issues that may hinder the tracker's performance: the aspect ratio of target objects in the templates can vary, some target objects may lack clear semantic meaning, and certain target objects may be indistinguishable from the background. Token-type embedding helps mitigate these issues by explicitly annotating foreground and background regions within the template. 

\textit{Positional embedding.} Positional embeddings~\cite{dosovitskiy2020image,huo2018learning} typically function for fixed resolutions. However, one-stream trackers often use image pairs of different sizes as inputs. According to the parameter-efficient fine-tuning (PEFT) design principle, sharing the positional embedding across these two images is crucial~\cite{yang2022exploiting}. Strategies for effectively reusing standard 1D absolute positional embeddings for varying input sizes are explored. Both strategies consider the original 1D absolute embedding within a 2D context. Let \(Q\) denote the original 1D absolute positional embedding; the 2D representation of \(Q\) can be expressed as \(Q_{2d} = [q_{1,1},...,q_{1,w}; q_{2,1},...,q_{2,w}; q_{h,1},...,q_{h,w}]\). For convenience, it is assumed that the resolution of the search region image matches the native resolution of the ViT model, allowing \(Q_{2d}\) to be directly applied to search region tokens, i.e., the search region positional embedding \(Q_x = Q_{2d}\). 

\textbf{MLP-only Head Network.} To mitigate potential inductive biases inherent in convolutional heads, a multi-layer perceptron (MLP) only head is employed. The head network generates a redundant set of bounding boxes, each accompanied by a classification score. These classification scores rank the bounding boxes, with the highest score determining the final output. The head network is structured into two branches: one for classification and another for bounding box regression, each consisting of a three-layer MLP. Both branches process the feature map of the search area \(\mathcal{S}\) outputted from the Transformer encoder. The head network adopts a center-based, anchor-free style ~\cite{tian2019fcos} to accelerate training coverage and enhance tracking performance.

\section{Experiments}

\textbf{Dataset.} We use a fine-tuned evaluation method. The experimental evaluation utilizes three datasets: Perception Test, LaSOT~\cite{fan2019lasot}, and GOT-10k~\cite{huang2019got}. The Perception Test dataset is divided into training, validation, and test subsets, with the training and validation sets used to fine-tune the model, and the test set reserved for performance evaluation. See details about the Perception Test dataset in Table \ref{tab:1}. Additionally, the LaSOT and GOT-10k datasets are exclusively employed as training data.

\begin{table}[h]
  \caption{Perception Test dataset}
  \label{tab:1}
  \centering
  \begin{tabular}{ccc}
    \toprule
    split&videos&object tracks\\
    \midrule
    Train & 2184 & 35373\\
    Validation & 1000 & 16501\\
    Test & 1000 & 16339\\
  \bottomrule
\end{tabular}
\end{table}

\textbf{Metric.} The evaluation metric is the average Intersection over Union (IoU). It is calculated as the average IoU between the predicted bounding boxes and the ground truth bounding boxes for each tracked object.

\textbf{Comparison Methods Result.} To select the most suitable model for the foundational component of the solution, comparative experiments are conducted on the test set. Table \ref{tab:2} illustrates the evaluation of four model approaches: Dummy Static, SiamFC, SAM2, and LoRAT.

\begin{table}[h]
  \caption{Experimental results of different methods}
  \label{tab:2}
  \centering
  \begin{tabular}{cc}
    \toprule
    Method&average IoU \\
    \midrule
    Dummy static & 0.640\\
    SiamFC & 0.658\\
    SAM2 & 0.693\\
    LoRAT & 0.802\\
  \bottomrule
\end{tabular}
\end{table}

Through experiments on the test set, we can see that the LoRAT method is the best on the indicator of IoU. According to this result, we select the LoRAT model as the basic model for our solution. Subsequent experiments are also based on this model.

\textbf{Ablation Study.} To analyze the Contribution of each component to LoRAT, we conduct more ablation studies on the test set of the competition. The average Intersection
over Union after adding components is demonstrated in Table \ref{tab:3}. We implement the Alpha-Refine method to post-process the bounding box output generated by the model. Although Alpha-Refine aims to improve tracking performance through accurate bounding box estimation, its performance is not good enough.

\begin{table}[h]
  \caption{Ablation experiment. The B in B-224 stands for ViT-B, and 224 means the search region size is set to [224 × 224]. The L in L-378 stands for ViT-L, and 378 means the search region size is set to [378 × 378].}
  \label{tab:3}
  \centering
  \begin{tabular}{cc}
    \toprule
    Method&average IoU\\
    \midrule
    LoRAT-B-224 & 0.802\\
    LoRAT-L-378 & 0.810 \\
    LoRAT-L-378+Alpha-Refine & 0.785\\
    LoRAT-L-378+LaSOT+GOT-10k & 0.813\\
  \bottomrule
\end{tabular}
\end{table}

\section{Conclusion}
This report presents a comprehensive evaluation of single-object tracking methods, highlighting the selection and application of the LoRAT model as the core tracking algorithm. By leveraging diverse datasets, including Perception Test, LaSOT, and GOT-10k, and employing advanced techniques such as interpolation-based positional embedding adaptation, the study demonstrates significant improvements in tracking performance. Despite the limited impact of the Alpha-Refine method, the overall approach effectively addresses the challenges of single-object tracking, leading to a top ranking in the final test phase.

{\small
\bibliographystyle{unsrt}
\bibliography{main}
}
\end{document}